\documentclass{article}

\usepackage{microtype}
\usepackage{graphicx}
\usepackage{subfigure}
\usepackage{booktabs}
\usepackage{threeparttable}
\usepackage{makecell}
\usepackage{amsmath,amsfonts,bm,amssymb,mathtools}
\def\1{\bm{1}}

\def\rvu{{\mathbf{i}}}

\def\rvn{{\mathbf{n}}}

\def\rvu{{\mathbf{u}}}
\def\rvv{{\mathbf{v}}}
\def\rvw{{\mathbf{w}}}

\def\vtheta{{\bm{\theta}}}
\def\vSigma{{\bm{\Sigma}}}

\def\vphi{{\bm{\phi}}}

\def\vm{{\bm{m}}}

\def\mC{{\bm{C}}}

\def\mS{{\bm{S}}}

\DeclareMathAlphabet{\mathsfit}{\encodingdefault}{\sfdefault}{m}{sl}
\SetMathAlphabet{\mathsfit}{bold}{\encodingdefault}{\sfdefault}{bx}{n}

\def\gL{{\mathcal{L}}}

\newcommand{\E}{\mathbb{E}}

\newcommand{\KL}{\mathsf{D} _{\mathrm{KL}}}
\DeclarePairedDelimiterX{\infdivx}[2]{(}{)}{%
  #1\,\delimsize\|\,#2%
}

\usepackage{physics}
\newcommand{\Tsf}{\mathsf{T}}

\newcommand{\pnoise}{p_{\rvn_i}\kern-1pt}
\newcommand{\g}{\,\vert\,}
\usepackage{numprint}

\usepackage{hyperref}
\usepackage{cleveref}

\usepackage[accepted]{icml2020}

\icmltitlerunning{Density Deconvolution with Normalizing Flows}

\begin{document}

\twocolumn[
\icmltitle{Density Deconvolution with Normalizing Flows}

\icmlsetsymbol{equal}{*}

\begin{icmlauthorlist}
\icmlauthor{Tim Dockhorn}{equal,watvec}
\icmlauthor{James A. Ritchie}{equal,ed}
\icmlauthor{Yaoliang Yu}{watvec}
\icmlauthor{Iain Murray}{ed}
\end{icmlauthorlist}

\icmlaffiliation{watvec}{University of Waterloo and Vector Institute, Canada}
\icmlaffiliation{ed}{School of Informatics, University of Edinburgh, Edinburgh, United Kingdom}

\icmlcorrespondingauthor{Tim Dockhorn}{tim.dockhorn@uwaterloo.ca}
\icmlcorrespondingauthor{James A. Ritchie}{james.ritchie@ed.ac.uk}

\icmlkeywords{Machine Learning, Normalizing Flow, Deconvolution}

\vskip 0.3in
]

\printAffiliationsAndNotice{\icmlEqualContribution}

\begin{abstract}
Density deconvolution is the task of estimating a probability density function given only noise-corrupted samples.
We can fit a Gaussian mixture model to the underlying density by maximum likelihood if the noise is normally distributed, but would like to exploit the superior density estimation performance of normalizing flows and allow for arbitrary noise distributions. Since both adjustments lead to an intractable likelihood, we resort to amortized variational inference. We demonstrate some problems involved in this approach, however, experiments on real data
demonstrate that flows can already out-perform Gaussian mixtures for density deconvolution.
\end{abstract}
\section{Introduction}
Density estimation is the fundamental statistical task of estimating the density of a distribution given a finite set of measurements.
However, in many scientific fields~\citep[see examples in][]{carroll2006measurement}, one only has access to a noise-corrupted set of measurements. Given knowledge of the statistics of the noise, \emph{density deconvolution} methods attempt to recover the density function of the unobserved noise-free samples rather than the noisy measurements.

In this work we consider the problem of additive noise, where observed samples $\{\rvw_i\}_{i=1}^m$ are produced by adding independent noise to unobserved values $\{\rvv_i\}_{i=1}^m$,
\begin{align}
    \rvw_i = \rvv_i + \rvn_i.
\end{align}
We also assume that the density function of the noise $\pnoise(\rvn_i)$ is perfectly known for every observation. The density function of observations $\rvw_i$ is then a convolution
\begin{align}
\label{eqn:convolution}
   p(\rvw_i) = \int_\rvv \pnoise(\rvw_i - \rvv)\, p(\rvv) \, d\rvv.
\end{align}
When the noise distribution is constant, we could estimate the density of the observations $\rvw$ with any density estimator and then solve~\Cref{eqn:convolution}, e.g.\ with a kernel density estimator using Fourier transforms~\citep[e.g.,][]{liu1989consistent, carroll1988optimal, fan1991optimal, devroye1989consistent} or wavelet decompositions~\citep{pensky1999adaptive}. 

When the noise distribution is different for each observation, we only have one sample from each convolved density. This \emph{extreme deconvolution} setting \citep{bovy2011extreme} has previously been tackled by fitting a Gaussian Mixture Model (GMM) to the underlying density $p(\rvv)$. When the noise distributions are all Gaussian, the marginal likelihood $p(\rvw_i)$ is tractable, and the GMM can be fitted by Expectation-Maximisation \citep[EM,][]{bovy2011extreme} or Stochastic Gradient Descent \citep[SGD,][]{ritchie2019scalable}. 

Given enough mixture components, any density function can be approximated arbitrarily closely using GMMs. In practice, however, other representations of densities can be easier to fit, and often generalize better.
There is growing interest in normalizing flows~\citep{tabak2010density, tabak2013family}, a class of methods that transform a simple source density into a complex target density. Normalizing flows are an efficient alternative to GMMs~\citep[e.g.,][]{rezende2015variational}, providing both good scalability and high expressivity, and have shown promise in applications similar to density deconvolution~\citep{cranmer2019modeling}.\looseness=-1

In this work, we model the underlying density $p(\rvv)$ with a normalizing flow. The marginal likelihood $p(\rvw_i)$ is intractable, so we resort to approximate inference.
We use amortized variational inference (\Cref{sec:methods}), closely following
Variational Auto-Encoders \citep[VAEs,][]{kingma2014auto,rezende2014stochastic}. Unlike for VAEs, in our framework, the model between the latent $\rvv$ and observed $\rvw$ vectors is fixed.

In this proof of concept, we use a fixed Gaussian noise distribution, but our approach would also allow us to use arbitrary and varying noise distributions, as found in realistic applications \citep[e.g.,][]{anderson2018}.
In a setting well-suited to the existing Gaussian mixture approach, we find that
fitting flows is harder (\Cref{sec:mogexperiments}), possibly motivating further work on approximate
inference in this setting. Nevertheless, on real data, we demonstrate that flows can already
outperform GMMs for density deconvolution (\Cref{sec:uciexperiments}).

\section{Methods}
\label{sec:methods}
We take a variational approach~\citep{jordan1999introduction} to density deconvolution. Introducing an approximate posterior $q_\vphi(\rvv)$ gives a lower bound to the log-marginal likelihood
\begin{align}
    &\log p(\rvw_i) = \log \int_\rvv \pnoise(\rvw_i \!-\! \rvv)\, p_\vtheta(\rvv) \, d\rvv \\
    &= \log \int_\rvv \pnoise(\rvw_i \!-\! \rvv)\, p_\vtheta(\rvv)\, \frac{q_\vphi(\rvv)}{q_\vphi(\rvv)} \, d\rvv \\
    &\geq \E_{q}[\log \pnoise(\rvw_i \!-\! \rvv)] - \KL\infdivx{q_\vphi(\rvv)}{p_\vtheta(\rvv)} = \gL,
\end{align}
where $\gL$ is the evidence lower bound (ELBO, see \Cref{apx:elbo}). Our approximate posterior $q_\vphi(\rvv)$, a \emph{recognition network}, represents beliefs about an underlying value~$\rvv$ given an observation $\rvw$ and the parameters of the noise.

The ELBO gives a unified objective for both the parameters~$\vtheta$ of the model and the parameters~$\vphi$ of the recognition network. Stochastic gradient descent only needs unbiased estimates of the ELBO, which we obtain by Monte Carlo
\begin{align}
    \gL \approx \gL(K) = \frac{1}{K}\sum_{k=1}^K \log\left[ \frac{\pnoise(\rvw_i - \rvv_k)\, p_\vtheta(\rvv_k)}{ q_\vphi(\rvv_k)} \right],
    \label{eqn:varloss}
\end{align}
where $K$ Monte Carlo samples are simulated $\rvv_k \sim q_\vphi(\rvv)$.

Our variational approach follows that of Variational Auto-Encoders \citep[VAEs,][]{kingma2014auto, rezende2014stochastic}, which provide a framework for amortized variational inference in graphical models. The focus of VAEs, however, is usually to build generative models matching the observations $\rvw$. In contrast, our main target is estimating an underlying density function $p(\rvv)$. For certain applications, e.g.\ denoising a noisy measurement, we may also be interested in the approximate posterior $q_{\vphi}(\rvv)$. Another difference between our method and VAEs is that we do not learn a likelihood model between the latent variables $\rvv$ and the observations $\rvw$. Instead, our ``likelihood model" is fully-characterized by the problem itself as $\pnoise(\rvw_i - \rvv)$. 

We model both $p_\vtheta(\rvv)$ and $q_\vphi(\rvv)$ as normalizing flows. Normalizing flows model probability density functions by transforming a source density $\pi(\rvu)$ into a target density $\hat{\pi}(\rvv)$ using an invertible, differentiable transformation $\Tsf$
\begin{align}
   \rvv = \Tsf(\rvu).
\end{align}
The density of $\rvv$ can be computed using the change-of-variable formula
\begin{align}
    \hat{\pi}(\rvv) = \frac{\pi \left(\Tsf^{-1} (\rvv) \right)}{\abs{\det \frac{\partial \Tsf}{\partial \rvu}\left(\Tsf^{-1} (\rvv) \right)}}
\end{align}
In particular, we model $p_\vtheta(\rvv)$ and $q_\vphi(\rvv)$ with autoregressive flows.
For $p_\vtheta(\rvv)$ we use a Masked Autoregressive Flow \citep[MAF,][]{papamakarios2017masked}, where a single neural network pass can compute $\Tsf^{-1}(\rvv)$, and therefore the densities $\hat{\pi}(\rvv)$ required during training.
Inverting the network, to generate samples, requires $D$ neural network passes for $D$-dimensional data.
For $q_\vphi(\rvv)$ we use the same network architecture to represent $\Tsf(\rvu)$, corresponding to an Inverse Autoregressive Flow \citep[IAF,][]{kingma2016improved}. During training, this choice gives fast one-pass generation of samples with their densities.
For an extensive review on normalizing flows, we refer the reader to~\citet{papamakarios2019normalizing}.
\looseness=-1

\section{Related Work}
\label{sec:related}
\textbf{Importance weighting:} The Importance Weighted Autoencoder~\citep{burda2015importance} has the same architecture as the standard VAE, however, it is trained on a lower bound that is tighter than the standard ELBO\@. Applying this idea to our model results in the following lower bound:
\begin{align}
    \log p(\rvw_i) &\geq \log \left[ \frac{1}{K} \sum_{k=1}^{K} \frac{\pnoise(\rvw_i - \rvv_k)\, p_\vtheta(\rvv_k)}{q_\vphi(\rvv_k)} \right] \\[0.1in]
    &= \gL_{\rm{IW}}(K).
\end{align}
It can be shown~\citep{cremer2017reinterpreting} that, in expectation, $\gL_{\rm{IW}}(K)$ is equivalent to $\gL(K)$ with an implicit, more expressive approximate posterior. A theoretical advantage of $\gL_{\rm{IW}}(K)$ over $\gL(K)$ is that the former is consistent \citep[under some mild boundedness assumptions,][Theorem~1]{burda2015importance}, i.e., $\lim_{K \to \infty} \gL_{\rm{IW}}(K) = \log p(\rvw_i)$.

In fact, it is possible to construct an unbiased estimator of $\log p(\rvw_i)$ using $\gL_{\rm{IW}}(K)$ with finite $K$ \citep{Luo2020SUMO}, in combination with a Russian Roulette Estimator \citep{kahn1955use}. A drawback of this approach is that there is no guarantee that the variance of the estimator is finite.

\textbf{Inference suboptimality:} When using the ELBO, or $\gL_{\rm{IW}}(K)$ with finite $K$, we only have a bound on the marginal log-likelihood $\log p(\rvw_i)$. This bound is loose when the approximate posterior is incorrect, which happens either because the form of the posterior cannot be represented, or because the recognition network does not produce good variational parameters for all data points.
Both issues can be overcome by choosing the approximate posterior from an expressive variational family
~\citep{cremer2018inference}, which is why we use a flow.

\textbf{Expressive priors for representation learning:} 
The standard VAE has a fixed prior, usually a multivariate standard normal distribution, but VAEs with more expressive priors have been proposed. Expressive priors are particularly useful when the distribution of the latent variables is used for representation learning.

A simple generalization for the prior is a learnable GMM~\citep[e.g][]{nalisnick2016approximate, dilokthanakul2016deep}, which in our context would result in the existing extreme deconvolution model \citep{bovy2011extreme}, with no need for variational inference.
Another approach is to model the prior with a collection of categorical distributions~\citep[e.g.][]{van2017neural}, which would be appropriate if an observation is well-modeled as a composition of prototype sources.
We use MAF, because for the applications we have in mind \citep[e.g.,][]{anderson2018}, we want to use a flexible, continuous prior representation.
VAEs have also used autoregressive flow priors before \citep{chen2017variational}.

\section{Experiments}
\label{sec:experiments}
In this section, we compare our method to a baseline of GMMs fitted with the Extreme Deconvolution (XD) model \citep{bovy2011extreme},
as by \citet{ritchie2019scalable}.
\subsection{Mixture of Gaussians}
\label{sec:mogexperiments}
\begin{figure}
    \vskip -0.12in
    \includegraphics[width=\linewidth]{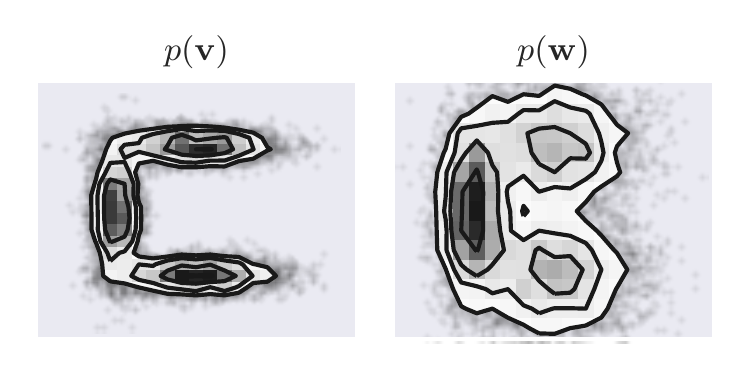}
    \vskip -0.2in
    \caption{2D histograms of training data for synthetic experiments. 
    Contour lines are estimated 0.5/1/1.5/2-$\sigma$ levels, with samples in the tails plotted directly.
    Left: Latent data sampled from a mixture of Gaussians. Right: Observed data created by adding noise to samples from $p(\rvv)$.}
    \label{fig:toy_data}
\end{figure}

In this synthetic task,
the target underlying density is a mixture of Gaussians. We fit the models from observations that include additional noise from a Gaussian with fixed covariance.
Figure~\ref{fig:toy_data} shows density plots of both the latent data $\rvv$ and the observed data $\rvw$.

The exact posterior for a latent datapoint given a noisy observation
is itself a mixture of Gaussians, and for some observations the components may be highly isolated.
We have deliberately picked an example where the prior $p(\rvv)$ should be reasonably easy for a flow to model, but the posterior $p(\rvv \g \rvw)$ may cause issues for flows, as they are known to have trouble fitting mixture of Gaussians when the components are well-separated~\citep[e.g.,][]{jaini2019sum}.
Full experimental details are reported in Appendix~\ref{apx:gmm_exp}

Table~\ref{tab:gmm_results} reports test average negative log-likelihood on both $\rvv$ and $\rvw$, referred to as $\log p(\rvv)$ and $\log p(\rvw)$, respectively. We estimate $\log p(\rvw)$ using $\gL_{\rm{IW}}(100)$.
The GMM, the true model class, has the best results for both $\rvv$ and $\rvw$.
The flows give quite close estimates for $\log p(\rvw)$ when $K>1$ for both $\gL$ and $\gL_{\rm{IW}}$, but show very high variance in their estimates of $\log p(\rvv)$ relative to the variance for $\log p(\rvw)$.

\begin{figure*}[ht]
    \vskip -0.1in
    \includegraphics[width=\linewidth]{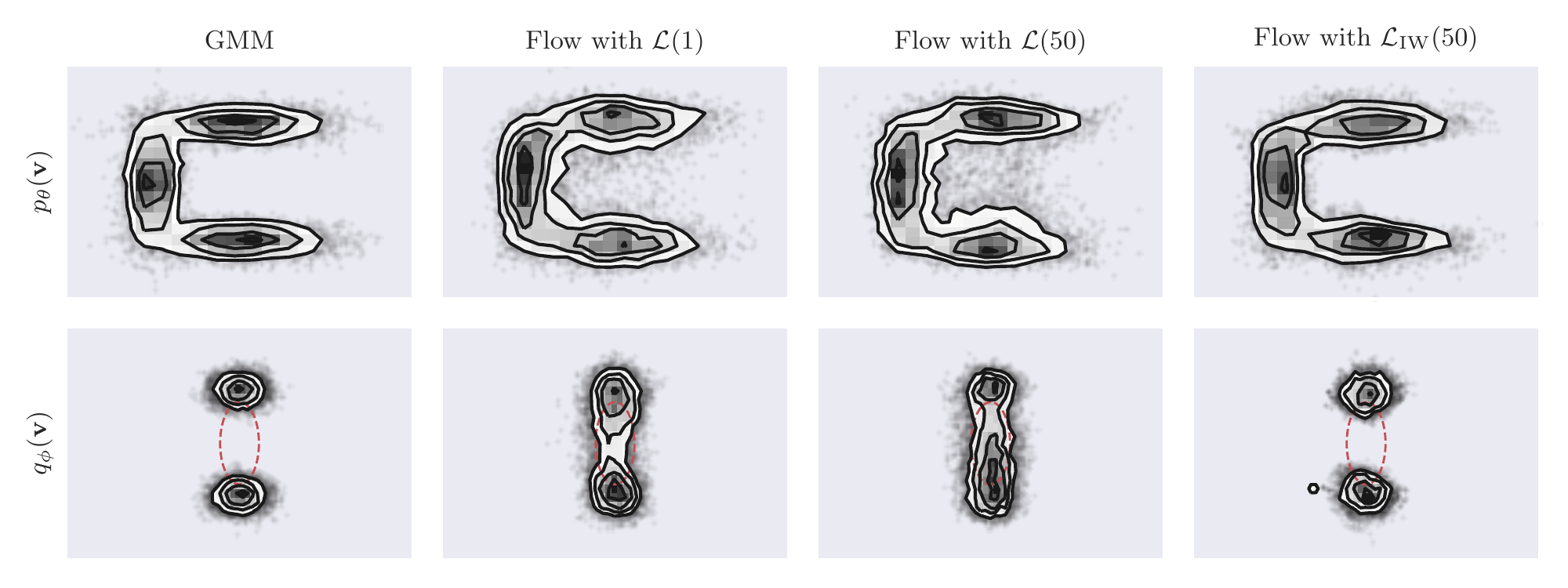}
    \vskip -0.2in
    \caption{Density plots for fitted models using the same representation as Figure~\ref{fig:toy_data}. The top row shows samples from the prior, whilst the bottom row shows samples from the corresponding posterior for a given noisy test point. The red dashed ellipse shows the 1-$\sigma$ level of the Gaussian noise around the test point.}
    \label{fig:toy_models}
\end{figure*}

The top row of Figure~\ref{fig:toy_models} shows example density plots using samples from the priors of our fitted models.
The fitted GMM has matched the ground truth GMM closely.
The flows recover the broad shape of the ground truth model, but those trained with $\gL(1)$ and $\gL(50)$ put too much mass in the centre.
The flow trained with $\gL_{\rm{IW}}(50)$ matched the Gaussian mixture model on the run shown, but the results had high variance, and other runs do not look as good.

The bottom row of Figure~\ref{fig:toy_models} shows example posteriors for the fitted models.
The exact posterior for the GMM has two isolated modes,
and is a close match to the ground truth posterior.
The approximate posteriors for the flows trained with $\gL(1)$ and $\gL(50)$ are not good matches to the GMM posterior, as neither have isolated modes, but are reasonably consistent with their corresponding priors.
The approximate posterior for the flow trained with $\gL_{\rm{IW}}(50)$ uses samples drawn from $q_{\vphi}$ with sampling-importance-resampling~\citep[e.g.,][]{rubin1988using}. The resampling reflects this recognition network's role as an adaptive proposal distribution under the $\gL_{\rm{IW}}$ objective rather than a direct approximation to the posterior~\cite{cremer2017reinterpreting}.
This reweighted approximation is a much better match to the ground truth posterior, but as with the prior, the variance across training runs was high, and other examples are qualitatively worse.

\begin{table}
    \centering
    \footnotesize
    \begin{tabular}{l l c c}
        \toprule
        Method & K & $-\log p(\rvv)$ & $-\log p(\rvw)$ \\
        \midrule
        XD-GMM & -- & $2.667 \pm 0.000$ & $3.600 \pm 0.000$ \\
        \midrule
        Flow ($\gL$) & 1%
        & $2.897 \pm 0.046$ &  $3.609 \pm 0.002$ \\
        & 10 &  $3.015 \pm 0.228$ &  $3.607 \pm 0.002$ \\
        & 25 &  $2.858 \pm 0.077$ &  $3.606 \pm 0.004$ \\
        & 50 &  $2.913 \pm 0.174$ &  $3.605 \pm 0.001$ \\
        \midrule
        Flow ($\gL_{\rm{IW}}$) & 10 & $2.854 \pm 0.083$ &  $3.604 \pm 0.002$ \\
        & 25 &  $2.871 \pm 0.045$ &  $3.604 \pm 0.001$ \\
        & 50 &  $3.070 \pm 0.499$ &  $3.603 \pm 0.001$ \\
        
        \bottomrule
    \end{tabular}
    \caption{Test average negative log-likelihood for the Gaussian mixture toy dataset. Average over five runs with standard deviation.}
    \label{tab:gmm_results}
\end{table}

To establish whether the inability of our procedure to consistently recover the correct ground truth model is a problem with the prior flow itself, the approximate posterior, the interaction of both, or the training objectives, we tried various methods of training each part separately.
All results for these experiments are summarized in Table~\ref{tab:pretraining_results}.
Additional density plots are also available in Appendix~\ref{apx:plots}.

The flow $p_{\vtheta}(\rvv)$ was pretrained directly on the noise free samples underlying the training set via maximum likelihood.
The test average negative log-likelihood on $\rvv$ is much closer to the GMM\@. Therefore, while model mismatch is a slight disadvantage for the flows here, it is not entirely responsible for their worse results.
Similarly, we trained the recognition network directly on noisy samples from the training set, paired with samples from the exact ground truth posterior.
When combined with the pretrained prior, the test likelihood on $\rvw$ was much closer to that of the GMM, suggesting that the flows can represent useful posteriors.

We then fitted the models with the $\gL(50)$ objective, but initialized with the pretrained prior and posterior. 
The variational objective was significantly improved by moving to a model with similar $\log p(\rvw)$ as the GMM, however, doing so made $\log p(\rvv)$ worse.
Despite using fairly flexible flows, the variational bound is not tight, and biases us towards worse prior models.

Finally we tried fitting a GMM with the $\gL(50)$ objective rather than by maximizing the log-likelihood directly, using both samples from the exact posterior and samples from the conditional flow approximate posterior.
When using exact samples, the variational bound is tight, but we experience the noisier gradients of variational fitting: the GMM still recovers the same result as before.
However, using posterior samples from the flow causes a similar bias to before, showing that the flows are not learning good enough posteriors for variational inference to be accurate.

\begin{table}
    \centering
    \footnotesize
    \begin{tabular}{l c c}
        \toprule
        Model & $-\log p(\rvv)$ & $-\log p(\rvw)$ \\
        \midrule
        Pretrained flows & $2.675 \pm 0.017$ & $3.601 \pm 0.002$ \\
        After variational fitting & $2.729 \pm 0.035$ & $3.602 \pm 0.001$ \\
        \midrule
        \makecell[l]{GMM, flow posterior} & $2.731 \pm 0.008$ & $3.602 \pm 0.000$ \\
        \makecell[l]{GMM, exact posterior} & $ 2.666 \pm 0.001$ & $3.600 \pm 0.000$ \\
        \bottomrule
    \end{tabular}
    \caption{Test average negative log-likelihood for the additional experiments. Average over five runs with standard deviation.}
    \label{tab:pretraining_results}
\end{table}

\subsection{UCI datasets}
\label{sec:uciexperiments}
We now compare the two methods on two small UCI datasets~\citep{dua2017machine}
that are
difficult to fit with GMMs \citep{uria2013rnade}. We discarded discrete-valued attributes and normalized the data. The datasets are then split into training and testing sets; 90\% are used for training and 10\% are used for testing. We subsequently add noise from a zero-mean independent normal distribution with diagonal covariance matrix $\vSigma_{ii} \!=\! 0.1$ to each noise-free training point to generate the observations~$\rvw_i$. 

For our method, we use the objective $\gL_{\rm{IW}}(50)$ as it yielded the best test average negative log-likelihood in the previous section. We note, however, that this might not be the best choice as the high-variance pattern for $\log p(\rvv)$~(see \Cref{tab:gmm_results}) might persist.
The test results are reported in~\Cref{tab:results_uci}; full experimental details can be found in~\Cref{app:uci}.
Flows outperformed GMMs in both cases.

As an ablation, we tried fitting conventional flows to the noisy observations $\rvw_i$, without correcting for the noise in any way
(\Cref{tab:ablation_study}). These flows beat the GMMs on both datasets, showing the importance of using good representations, however, the results are still significantly worse than flows with deconvolution (right column \Cref{tab:results_uci}).
\begin{table}
    \centering
    \footnotesize
    \begin{tabular}{l c c}
        \toprule
        &\multicolumn{2}{c}{$-\log p(\rvv)$} \\ \cmidrule(lr){2-3}
        Dataset &XD-GMM &Flow $\gL_{\rm{IW}}(50)$ \\
        \midrule
        White wine &$9.903 \pm 0.112$ &$8.685 \pm 0.082$\\
        Red wine &$8.775 \pm 0.152$ & $8.083 \pm 0.128$\\
        \bottomrule
    \end{tabular}
    \caption{Test average negative log-likelihood for two small UCI datasets. Average over five runs with standard deviation.
    }
    \label{tab:results_uci}
\end{table}
\section{Conclusion}
In this preliminary work, we have outlined an approach for density deconvolution using normalizing flows and arbitrary noise distributions by turning the deconvolution problem into an approximate inference problem. Our experiments on a toy setup have shown that variational inference with an inaccurate posterior
can prevent the model prior from learning the underlying noise-free density. For future work, we are planning to experiment with unbiased inference, e.g., using Markov chain Monte Carlo methods~\citep[e.g.,][]{glynn2014exact, Qiu2020Unbiased} or importance weighting in combination with the Russian Roulette Estimator~\citep[][]{Luo2020SUMO}. Nevertheless, we have already been able to demonstrate that normalizing flows fitted with our approach can beat GMMs for density deconvolution on (small) real-world datasets, which indicates that further research on how to fit normalizing flows in this context is worth pursuing.\looseness=-1 

\section*{Acknowledgements}
We are grateful to Artur Bekasov and Conor Durkan for discussions around choices of normalizing flow for this application and use of their flows library~\citep{NIPS2019_8969}.
Our experiments also made use of corner.py~\citep{corner}, Matplotlib~\citep{Hunter:2007}, NumPy~\citep{numpy}, Pandas~\citep{mckinney-proc-scipy-2010} and PyTorch~\citep{pytorch}.
This work was supported in part by the EPSRC Centre for Doctoral Training in Data Science, funded by the UK Engineering and Physical Sciences Research Council (grant EP/L016427/1) and the University of Edinburgh. 
Resources used in preparing this research were provided, in part, by the Province of Ontario, the Government of Canada through CIFAR, and companies sponsoring the Vector Institute. 
We gratefully acknowledge funding support from NSERC and the Canada CIFAR AI Chairs Program.
\bibliography{main}
\bibliographystyle{icml2020}

\appendix
\section{The Evidence Lower Bound}
\label{apx:elbo}
The KL divergence between the variational distribution and the posterior can be written as
\begin{align}
    &\KL\infdivx{q_\vphi(\rvv)}{p(\rvv \g \rvw_i)} \notag\\
    &\quad = \E_{q} \left[\log q_\vphi(\rvv) \!-\! \log p_\vtheta(\rvv \g \rvw_i) \right] \\
    &\quad =\E_{q} \left[\log q_\vphi(\rvv) \!-\! \log p_\vtheta(\rvv, \rvw_i) \right] + \log p_\vtheta(\rvw_i).
\end{align}
The joint distribution of $\rvv$ and $\rvw$ can be computed as
\begin{align}
    p_\vtheta(\rvv, \rvw_i) &= \int_\rvn p_\vtheta(\rvv, \rvw_i, \rvn_i) \, d\rvn_i \\
    &= \int_\rvn \delta(\rvw_i = \rvv + \rvn)\, p_\vtheta(\rvv)\, \pnoise(\rvn_i) \, d\rvn_i \\
    &= p_\vtheta(\rvv)\, \pnoise(\rvw_i - \rvv),
\end{align}
where $\delta(\cdot)$ is the Dirac delta distribution. Hence,
\begin{align}
    \log p_\vtheta(\rvw_i) = \hbox{}& \KL\infdivx{q_\vphi(\rvv)}{p_\vtheta(\rvv \g \rvw_i)} \notag\\
    & \quad+ \E_{q}[\log \pnoise(\rvw_i - \rvv)] \notag\\
    & \quad- \KL\infdivx{q_\vphi(\rvv)}{p_\vtheta(\rvv)} \\
    = \hbox{}& \KL\infdivx{q_\vphi(\rvv)}{p_\vtheta(\rvv \g \rvw_i)} + \gL.
\end{align}

\section{Details for Experiments}

All code used to run the experiments is available:\\ \url{https://github.com/bayesiains/density-deconvolution}

\subsection{Mixture of Three Gaussians}
\label{apx:gmm_exp}

Latent datapoints $\rvv_i$ were drawn from a mixture of 3 Gaussians, with equal mixture weights, means
\begin{equation}
    \vm_1 = \begin{bmatrix} -2 \!&\! 0\end{bmatrix}^T\kern-3pt,\,
    \vm_2 = \begin{bmatrix} 0  \!&\! -2\end{bmatrix}^T\kern-3pt,\,
    \vm_3 = \begin{bmatrix} 0  \!&\! 2\end{bmatrix}^T\kern-3pt,
\end{equation}
and covariances
\begin{equation}
    \mC_1 = \begin{bmatrix} 0.3^2 & 0 \\ 0 & 1 \end{bmatrix},\quad
    \mC_2 = \mC_3 = \begin{bmatrix} 1 & 0 \\ 0 & 0.3^2 \end{bmatrix}.
\end{equation}
Zero-mean Gaussian noise with covariance
\begin{equation}
    \mS = \begin{bmatrix} 0.1 & 0 \\ 0 & 1 \end{bmatrix},
\end{equation}
was added to each $\rvv_i$ to produce $\rvw_i$.
Training and test sets consisted of \numprint{50000} samples each, whilst the validation set had \numprint{12500} samples.

The prior $p_{\vtheta}(\rvv)$ was modelled with a standard normal base distribution with 5 layers of an affine Masked Autoregressive Flow (MAF) interspersed with linear transforms parameterized by an LU-decomposition and a random permutation matrix fixed at the start of training, following \citet{NIPS2019_8969}.
A residual network~\citep{he2016deep} was used within each MAF layer, with 2 pre-activation residual blocks~\citep{he2016identity}.
Each block used two dense layers with 128 hidden features each.
Masking of the residual blocks was done using the ResMADE architecture~\citep{pmlr-v97-durkan19a}.

The recognition network $q_{\vphi}(\rvv)$ used a setup adapted from~\citet{NIPS2019_8969}, where the same flow configuration as the prior modeled the inverse transformation, making it an Inverse Autoregressive Flow \citep[IAF,][]{kingma2016improved}.
Conditioning was done by concatenating $\rvw$ to a flattened Cholesky decomposition of the noise covariance S and applying a 2 block residual network to produce a 64-dimensional embedding vector.
This vector was then concatenated directly onto the input of the neural network in every IAF layer.
Whilst conditioning on the noise covariance S was not strictly necessary, because it was fixed for this experiment, we included it so that our implementation could handle the Extreme Deconvolution case where each observation $\rvw_i$ has its own associated noise covariance $\mS_i$.

We trained with Adam \citep{kingma2014adam}, with initial learning rate $0.0001$, other parameters set to defaults, a minibatch size of 512, and with dropout \citep{JMLR:v15:srivastava14a} probability~0.2.
We trained for 300 epochs, and reduced the learning rate by a factor of $0.8$ if there was no improvement in validation loss for 20 epochs.

\subsection{UCI datasets}
\label{app:uci}
\begin{table}
    \centering
    \footnotesize
    \begin{tabular}{l c}
        \toprule
        Dataset & $-\log p(\rvv)$ \\
        \midrule
        White wine &$9.544 \pm 0.184$ \\
        Red Wine &$8.611 \pm 0.254$\\
        \bottomrule
    \end{tabular}
    \caption{Test average negative log-likelihood for two small UCI datasets. Average over five runs with standard deviation. The flow model is trained directly on noisy-observations $\rvw_i$ using maximum likelihood learning.}
    \label{tab:ablation_study}
\end{table}
Since the datasets are relatively small, we tune the hyperparameters of the models using 5-fold cross-validation and grid search; the parameters of the grid search are reported in~\Cref{tab:grid_search}. Once the hyperparameter values had been determined, we trained the models using a tenth of the training data for early-stopping and measured their performance on the 10\% held-out test data. 

In contrast to the setup in ~\Cref{apx:gmm_exp}, we used simple dense layers rather than residual layers within each MAF layer. Masking of the dense layers was done using the standard MADE architecture~\citep{germain2015made}. A further difference is that we conditioned the recognition network $q_\vphi(\rvv)$ only on $\rvw$.

Training ran until no improvement in validation loss was observed for 30 epochs. For this experiment, we did not apply any Dropout. The minibatch size was chosen to be~100.
\begin{table}
    \centering
    \footnotesize
    \begin{tabular}{ll}
    \toprule
        Hyperparameters &Tested values \\
        \midrule
        Fixed learning rate & 0.001\textsuperscript{\textasteriskcentered\textdagger}, 0.0005, 0.0001 \\
        MAF layers ($p_\vtheta(\rvv)$) & 3\textsuperscript{\textasteriskcentered}, 4, 5\textsuperscript{\textdagger} \\
        MAF layers ($q_\vphi(\rvv)$) & 3, 4\textsuperscript{\textasteriskcentered}, 5\textsuperscript{\textdagger} \\
        MAF hidden features & 64, 128\textsuperscript{\textasteriskcentered\textdagger} \\
        MAF hidden blocks & 1\textsuperscript{\textasteriskcentered\textdagger}, 2 \\
        \midrule
        Fixed learning rate & 0.01\textsuperscript{\textasteriskcentered}, 0.005\textsuperscript{\textdagger}, 0.001 \\
        \# of mixture components & 20\textsuperscript{\textdagger}, 50, 100, 200, 300\textsuperscript{\textasteriskcentered} \\
        \bottomrule
    \end{tabular}
    \caption{Tested values for hyperparameter tuning on UCI datasets. The chosen values for each dataset are marked as: (\textasteriskcentered) White wine, (\textdagger) Red wine.}
    \label{tab:grid_search}
\end{table}

\section{Additional Plots}
\label{apx:plots}

\begin{figure}[h]
    \vskip -0.1in
    \includegraphics[width=\linewidth]{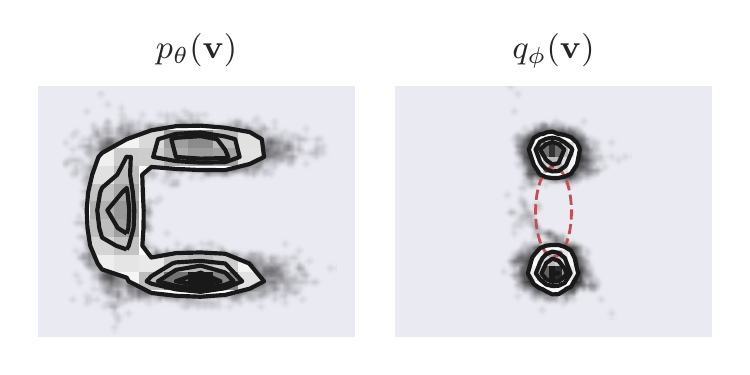}
    \vskip -0.2in
    \caption{Density plots of pretrained models. Left: Flow trained directly on noise-free samples. Right: Posterior flow trained as a conditional density estimator on pairs of noisy observations and samples from the exact posterior. The trail linking the posterior modes does not have a large penalty under our objective function.}
    \label{fig:add_0}
\end{figure}

\begin{figure}[h!]
    \vskip -0.1in
    \includegraphics[width=\linewidth]{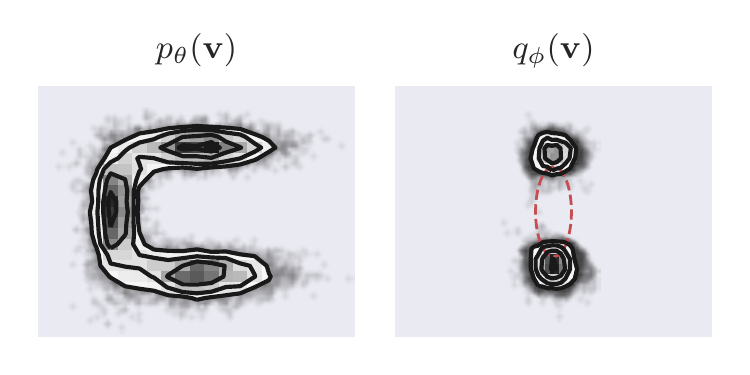}
    \vskip -0.2in
    \caption{Density plots of models initialized with pretrained flows, then trained jointly with $\gL(50)$. Left: Prior $p_{\vtheta}(\rvv)$. Right: Posterior $q_{\vphi}(\rvv)$ for a given test point. The trail between the posterior modes has been reduced, but is still present. The fitted prior density has gotten slightly worse (\Cref{tab:pretraining_results}).}
    \label{fig:add_1}
\end{figure}

\newpage

\begin{figure}[h]
    \vskip -0.1in
    \includegraphics[width=\linewidth]{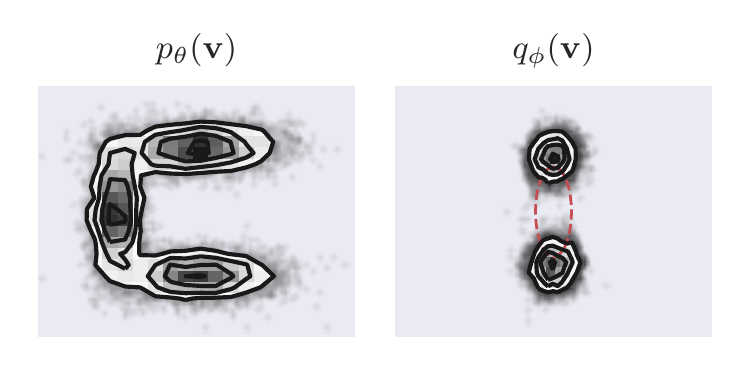}
    \vskip -0.2in
    \caption{Density plots of a GMM fitted with the $\gL(50)$ objective using a conditional-flow posterior. Left: Prior $p_{\vtheta}(\rvv)$. Right: Posterior $q_{\vphi}(\rvv)$ for a given test point. We are using the ground-truth model class, but the fit of the prior is not as good as when using the exact posterior. Importance weighted training may help remove the mass between the modes (\Cref{fig:toy_models}, right), but we did not get stable results (\Cref{tab:gmm_results}).}
    \label{fig:add_2}
\end{figure}

\begin{figure}[h]
    \vskip -0.1in
    \includegraphics[width=\linewidth]{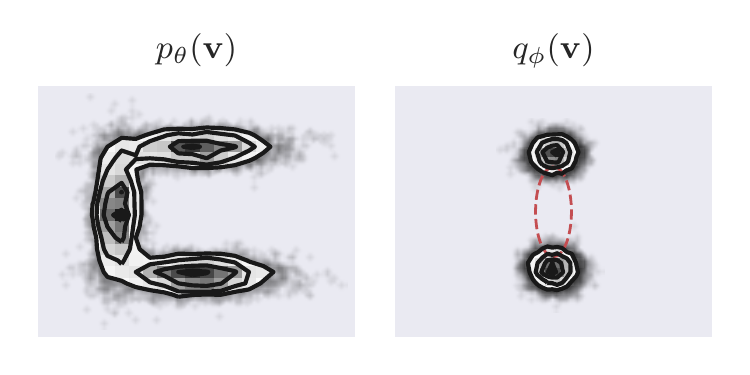}
    \vskip -0.2in
    \caption{Density plots of a GMM fitted with the $\gL(50)$ objective using the exact posterior. Left: Prior $p_{\vtheta}(\rvv)$. Right: Posterior $q_{\vphi}(\rvv)$ for a given test point.}
    \label{fig:add_3}
\end{figure}

\end{document}